\documentclass[11pt,a4paper]{article}
\usepackage[hyperref]{acl2020}
\usepackage{times}
\usepackage{latexsym}

\usepackage{microtype}

\aclfinalcopy % Uncomment this line for the final submission
\usepackage{tabularx} % extra features for tabular environment
\usepackage{array}
\newcolumntype{L}[1]{>{\raggedright\let\newline\\\arraybackslash\hspace{0pt}}m{#1}}
\newcolumntype{C}[1]{>{\centering\let\newline\\\arraybackslash\hspace{0pt}}m{#1}}
\newcolumntype{R}[1]{>{\raggedleft\let\newline\\\arraybackslash\hspace{0pt}}m{#1}}

\usepackage[hang,flushmargin]{footmisc} % footnote indent

\usepackage{amsmath}  % improve math presentation
\usepackage{graphicx} % takes care of graphic including machinery
\usepackage{mathtools}
\usepackage{amssymb}
\usepackage{booktabs} % For formal tables
\usepackage{multirow}
\usepackage{adjustbox}
\usepackage{makecell}
\usepackage{threeparttable}
\usepackage{subfigure}
\usepackage{overpic}
\usepackage{bm} % bm
\usepackage{soul}

\usepackage{balance} % balance references
\usepackage{booktabs} % professional table
\usepackage{threeparttable} % table footnote
\usepackage{enumitem} % enumerate

\newcommand{\RNum}[1]{\uppercase\expandafter{\romannumeral #1\relax}}

\newsavebox\CBox

\title{Semi-Supervised Dialogue Policy Learning via \\Stochastic Reward Estimation}

 \author{Xinting Huang,\textsuperscript{1}
Jianzhong Qi,\textsuperscript{1}
Yu Sun,\textsuperscript{2}
Rui Zhang\textsuperscript{1}\Thanks{ Rui Zhang is the corresponding author.}\\
\textsuperscript{1}{The University of Melbourne}, 
\textsuperscript{2}{Twitter Inc.}\\
\{xintingh@student., jianzhong.qi@, rui.zhang@\}unimelb.edu.au,
ysun@twitter.com}

\date{}

\begin{document}
\maketitle
\begin{abstract}
Dialogue policy optimization often obtains feedback until task completion in task-oriented dialogue systems. 
This is insufficient for training intermediate dialogue turns since supervision signals (or \emph{rewards}) are only provided at the end of dialogues.
To address this issue, reward learning has been introduced to learn from state-action pairs of an optimal policy to provide turn-by-turn rewards. 
This approach requires complete state-action annotations of human-to-human dialogues (i.e., expert demonstrations), which is labor intensive. 
To overcome this limitation, we propose a novel reward learning approach for semi-supervised policy learning.
The proposed approach learns a dynamics model as the reward function which models dialogue progress (i.e., state-action sequences) based on expert demonstrations, either with or without annotations.
The dynamics model computes rewards by predicting whether the dialogue progress is consistent with expert demonstrations.
We further propose to learn action embeddings for a better generalization of the reward function.
The proposed approach outperforms competitive policy learning baselines on MultiWOZ, a benchmark multi-domain dataset.

\end{abstract}

\section{Introduction}
Task-oriented dialogue systems complete tasks for users, such as making a restaurant reservation or finding attractions to visit, in multi-turn dialogues \cite{gao2018neural,sun2016contextual,sun2017collaborative}.
Dialogue policy is a critical component in both the conventional pipeline approach \cite{young2013pomdp} and recent end-to-end approaches \cite{zhao2019rethinking}.
It decides the next action that a dialogue system should take at each turn.
Considering its nature of sequential decision making, dialogue policy is usually learned via reinforcement learning \cite{su2015reward,peng2018deep,zhang2019budgeted}.
Specifically, dialogue policy is learned by maximizing accumulated rewards over interactions with an environment (i.e., actual users or a user simulator).
Handcrafted rewards are commonly used for policy learning in earlier work \cite{peng2018deep}, which assigns a small negative penalty at each turn and a large positive/negative reward when the task is successful/failed.
However, such reward setting does not provide sufficient supervision signals in each turn other than the last turn,
which causes the sparse reward issues and may result in poorly learned policies \cite{takanobu2019guided}.

\begin{table} [!tbp]
    \centering
\begin{threeparttable}
    \centering
\caption{ \small State Action Annotation and Utterance Example 
} \label{toy-example}
% \tiny
%\scriptsize
%\footnotesize
\small
% \normalsize  
\setlength\tabcolsep{3.0pt}
\begin{tabular}{L{1.05cm}|L{6.3cm}}
\toprule
% \multicolumn{2}{c}{\textbf{System utterances}} \\
\multirow{4}{*}{\makecell[l]{User\\Side} 
} & \textbf{Utterance} \\
\cmidrule{2-2}
& \textit{I would like moderate price range please.} \\ 
\cmidrule{2-2}
& \textbf{Dialogue State annotation} \\ 
\cmidrule{2-2}
& \texttt{Restaurant: \{food=modern european, price range=moderate\}}    \\
\midrule
\multirow{4}{*}{\makecell[l]{System\\Side} 
} & \textbf{Utterance} \\
\cmidrule{2-2}
&  \textit{I found de luca cucina and riverside brasserie. does either of them sound good for you?}  \\ 
\cmidrule{2-2}
& \textbf{System action annotation} \\ 
\cmidrule{2-2}
& \texttt{restaurant-inform:\{name=de luca cucina, name=riverside brasserie\}} \\
\bottomrule
\end{tabular}
% \begin{tablenotes}\footnotesize
% \item[*]
% \end{tablenotes}
\end{threeparttable}
\end{table}

To address this problem, reward function learning that relies on \emph{expert demonstrations} has been introduced \cite{takanobu2019guided,li2019dialogue}.
Specifically, state-action sequences generated by an optimal policy (i.e., expert demonstrations) are collected, and a reward function is learned to give high rewards to state-action pairs that better resemble the behaviors of the optimal policy.
In this way, turn-by-turn rewards estimated by the reward function can be provided to learn dialogue policy.
Obtaining expert demonstrations is critical to reward function learning.
Since it is impractical to assume that an optimal policy is always available, a common and reasonable approach is to treat the decision makings in human-human dialogues as optimal behaviors.
To accommodate the learning of reward function, human-human dialogues need to be annotated in the form of \emph{state-action pairs} from textual utterances.
Table \ref{toy-example} illustrates an example of human-human dialogue and its state-action annotation.
However, obtaining such annotations require extensive efforts and costs. 
Besides, a reward function based on state-action pair might cause an unstable policy learning, especially with a limited amount of annotated dialogues \cite{yang2018unsupervised}.

To address the above issues, we propose to learn dialogue policies in a semi-supervised setting where the system action of expert demonstrations only need to be partially annotated.
We propose to use an implicitly trained \emph{stochastic dynamics model} as the reward function to replace the conventional reward function that is restricted to state-action pairs.
Dynamics models describe sequential progress using a combination of stochastic and deterministic states in a latent space, which promotes an effective tracking and forecasting \cite{minderer2019unsupervised,sun2019stochastic,Wang2019EnhancingIS}.
In our scenario, we train the dynamics model to describe dialogue progress of expert demonstrations.
The main rationale is that the reward function should give high rewards to actions that lead to dialogue progress similar to those in expert demonstrations.
This is because dialogue progress at the early stage highly influences subsequent progress, and the latter directly determines whether the task can be completed.
Since the learning of dynamics model maps observations to latent states and further reason over the latent states, we are no longer restricted to fully annotated dialogues.
Using dynamics model as reward function also promotes a more stable policy learning.

Learning the dynamics model in the text space is, however, prone to compounding errors due to complexities and diversities of languages.
We tackle this challenge by learning the dynamics model in an \emph{action embedding} space that encodes the effect of system utterances on dialogue progress.
We achieve action embedding learning by incorporating an embedding function into a generative models framework for semi-supervised learning \cite{kingma2014semi}.
We observe that system utterances with comparable effects on dialogue progress will lead to similar state transitions \cite{huang2019mala}. 
Therefore, we formulate the generative model to describe the state transition process.
Using the generative model, we enrich the expert dialogues (either fully or partially annotated) with action embedding to learn the dynamics model.  
Moreover, we also consider the scenarios where both state and action annotations are absent in most expert dialogues, referred to as unlabeled dialogues. 
To expand the proposed approach to such scenarios, we further propose to model dialogue progress using action sequences and reformulate the generative model accordingly.

% Contribution bullet points
Our contributions are summarized as follows:
\begin{itemize}[topsep=0pt,leftmargin=*,noitemsep,wide=0pt]
\item 
To the best of our knowledge, we are the first to approach semi-supervised dialogue policy learning.
\item
We propose a novel reward estimation approach to dialogue policy learning which relives the requirements of extensive annotations and promotes a stable learning of dialogue policy.  
\item
We propose an action embedding learning technique to effectively train the reward estimator from either partially labeled or unlabeled dialogues.
\item
We conduct extensive experiments on the benchmark multi-domain dataset.
Results show that our approach consistently outperforms strong baselines coupled with semi-supervised learning techniques. 
\end{itemize}

\section{Preliminaries}
For task-oriented dialogues, a dialogue policy $\pi(a|s)$ decides an action $a \in \mathcal{A}$ based on the dialogue state $s \in \mathcal{S}$ at each turn, where $\mathcal{A}$ and $\mathcal{S}$ are the predefined sets of all actions and states, respectively.
Reinforcement learning is commonly applied to dialogue policy learning, where the dialogue policy model is trained to maximize accumulative rewards through interactions with environments (i.e., users):
\begin{equation}\label{policy-learning}
    \mathcal{L}_{} = -\mathbb{E}_{\tau_i \sim \pi}[r(\tau)]= -\mathbb{E}_{\tau_i \sim \pi}[\sum_{t}r(s_t, a_t)]
\end{equation}
where $\tau_i = \{(s_t, a_t)|0 \leq t \leq n_{\tau}\}$ represents a sampled dialogue, and $r(\tau_i)$ is the numerical rewards obtained in this dialogue. 
Instead of determining $r(\tau_i)$ via heuristics, recent reward learning approaches train a reward function $r_{\theta}$ to assign numerical rewards for each state-action pair.
The reward function is learned from expert demonstrations $D_{demo}$ that are dialogues sampled from an optimal policy in the form of state-action pairs. 
Adversarial learning is usually adopted to enforces higher rewards to state-action pairs from expert demonstrations and lower rewards to those sampled from the learning policy \cite{fu2017learning}:
\begin{equation}
    \mathcal{L}_{\text{}} = -\mathbb{E}_{\tau_j \sim D_{demo}}[r_{\theta}(\tau_j)] + \log \mathbb{E}_{\tau_i \sim \pi }(\dfrac{\exp{r_{\theta}(\tau_i)}}{q(\tau_i)})
\end{equation}
where $\pi$ is the current dialogue policy, and $q$ is the distribution of dialogues generated with $\pi$.
In this way, the dialogue policy and reward function are iteratively optimized, which requires great training efforts and might lead to unstable learning results \cite{yang2018unsupervised}.
Moreover, such a reward learning approach requires a complete dialogue state and system action annotation of expert demonstrations, which are expensive to obtain.

\section{Proposed Model}

\begin{figure}[t]
\centering
\includegraphics[width=7.6cm]{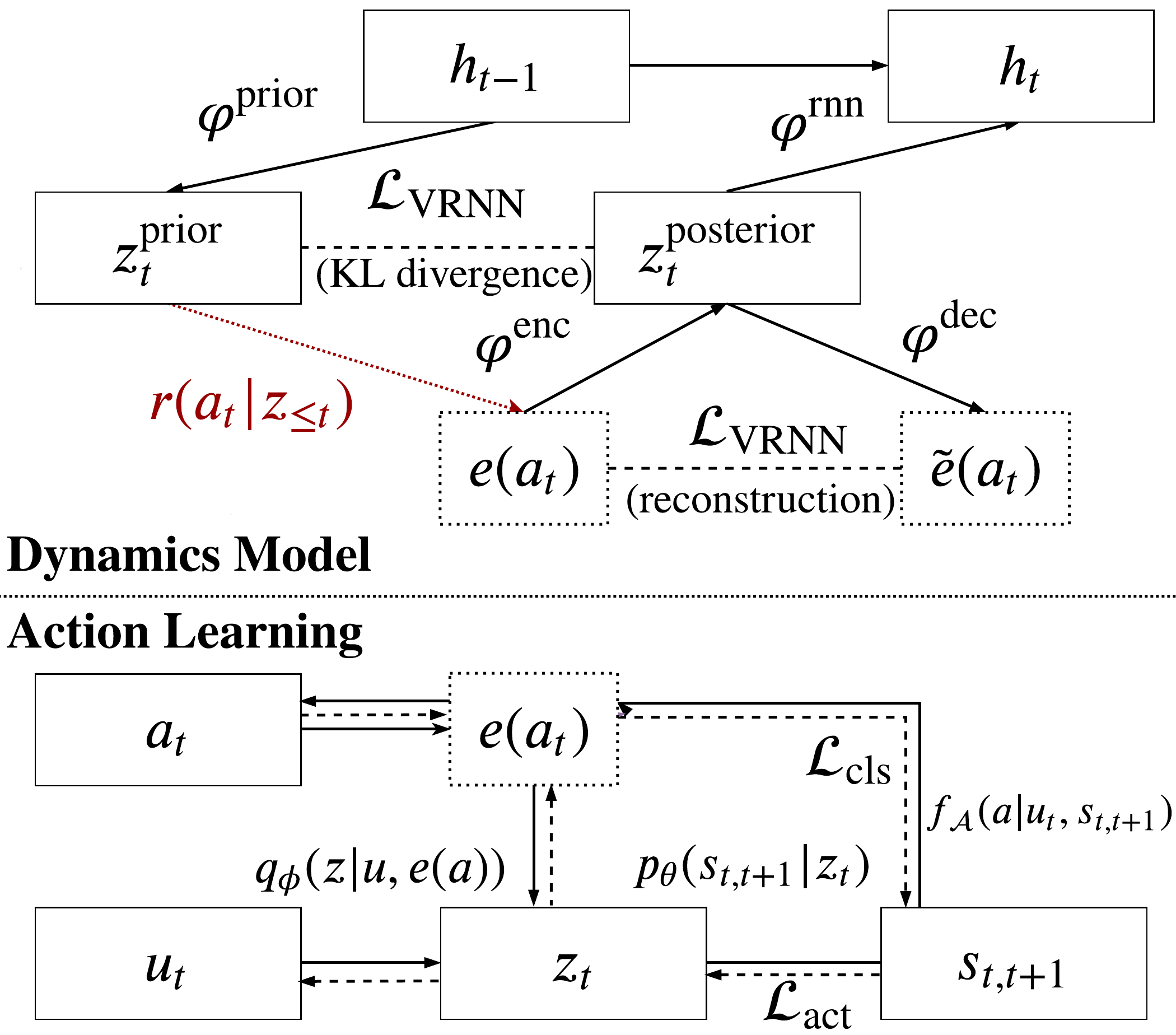}
\caption{
  Overall framework of the proposed approach
}
\label{overall-fig}
\end{figure}

\subsection{Overview}
We study the problem of \emph{semi-supervised} dialogue policy learning. 
Specifically, we consider the setting that expert demonstrations $D_{demo}$ consist of a small number of fully labeled dialogues $D_{\mathcal{F}} $ and partially labeled dialogues $D_{\mathcal{P}}$.
For each fully annotated dialogue $\tau_i$ in $D_{\mathcal{F}}$, complete annotations are available: $\tau_i = \{(s_t, a_t, u_t)|1 \leq t \leq n_{\tau} \}$, where $u_t$ is the system utterance at turn $t$.
Meanwhile, each partially labeled dialogue $\tau_j$ in $D_{\mathcal{P}}$ only has state annotations and system utterances: $\tau_j = \{(s_t, u_t)|1 \leq t \leq n_{\tau} \} $.

Figure \ref{overall-fig} illustrates the overall framework of the proposed approach. 
Rewards are estimated by a dynamics model that consumes action embeddings $e(a_t)$.
% design an objective by extending variational inference to 
% We design a 
Every action in the set $\mathcal{A}$ is mapped to a fix-length embedding via a learnable embedding function $f_E$.
To obtain the action embeddings for $D_{\mathcal{P}}$ which has no action annotations, we first predict the action via a prediction model $f_{\mathcal{A}}$ and then transform the predicted actions to embeddings.
To obtain effective action embeddings, we design a state-transition based objective to jointly optimize $f_E$ and $f_{\mathcal{A}}$ via variational inference (Sec. \ref{action-learning}). 
After obtaining the action embeddings, the dynamics model is learned by fitting the expert demonstrations enriched by action embeddings.  
Rewards are then estimated as the conditional probability of the action given the current dialogue progress encoded in latent states (Sec. \ref{reward-estimation}). 
We also extend the above approach to unlabeled dialogues where both state and action annotations are absent (Sec. \ref{expand-unlabel}).

\subsection{Action Learning via Generative Models} \label{action-learning}
% problem definition
We aim to learn the prediction model $f_{\mathcal{A}}$ and action embeddings using both $D_{\mathcal{F}}$ and $D_{\mathcal{P}}$.
% We formulate the action prediction model as a mapping from system utterances $x_t$ and its corresponding state transition $(s_{t-1}, s_t)$ to an action $a \in \mathcal{A}$: $f_{\mathcal{A}}(a|x_t, s_{t-1},s_t)$.
We formulate the action prediction model as $f_{\mathcal{A}}(a|u_t, s_{t},s_{t+1})$ which takes as input the system utterance $u_t$ and its corresponding state transition $(s_{t}, s_{t+1})$. 
% action embedding space: $\mathcal{E} \subseteq \mathbb{R}^{d}$, and an
We then introduce an mapping function: $f_{E}: \mathcal{A} \rightarrow \mathcal{E}$, where $\mathcal{E} \subseteq \mathbb{R}^{d}$ is the action embedding space later used for learning the dynamics model.
% where the action embedding space $\mathcal{E} \subseteq \mathbb{R}^{d}$ will be used for the learning of dynamics model. 
% We then introduce an action embedding space: $\mathcal{E} \subseteq \mathbb{R}^{d}$, where $d$ is the dimension of action embeddings, and an embedding-to-action mapping function: $f: \mathcal{E}\rightarrow \mathcal{A} $.

% \subsubsection*{Semi-Supervised Learning with generative model}
We train the prediction model by proposing a variational inference approach based on a semi-supervised variational autoencoder (Semi-VAE) \cite{kingma2014semi}.
% a variational inference approach which is built on semi-VAE framework
% treat system action as a hidden variable.
% We first recap the classic VAE framework for semi-supervised learning.
Semi-VAE describes the data generation process of feature-label pairs $\{({x_i}, y_i)|1 \leq i\leq N \}$ via latent variables ${z}$ as:
% construct a deep generative model to describe the data generation using a latent class variable $y$ in addition to a latent variable $z$:
\begin{equation}\label{generation-model}
   \log p({x}) = \log \sum_{y} \int_{z} p_{\theta}({x},z,y)dz  
\end{equation}
where $p_{\theta}$ is a generative model parameterised by $\theta$, and the class label $y$ is treated as a latent variables for unlabeled data. 
Since this log-likelihood in Eqn. \ref{generation-model} is intractable, its variational lower bound for unlabeled data is instead optimized as:
% variational inference \cite{kingma2014semi} is utilized to optimize the variational bound of Eqn. \ref{generation-model}:
% For data instances with labels, the objective is formulated as: 
% \begin{equation}
%     \log p(x, y) \geq \mathbb{E}_{q_{\phi}(z|x, y) } { [\log p_{\theta}(x|z, y)]}- KL(\cdot||\cdot)
% \end{equation}
% Similarly, the objective for unlabeled data is formulated based on such variational approximation:
\begin{equation}
\begin{split}
    \log p(x) \geq \mathbb{E}_{q_{\phi, \psi}(y, z|x)}[\log \frac{p_{\theta}(x, z, y)}{q_{\phi, \psi}(y, z|x)}] \\
    = \mathbb{E}_{q_{\psi}(y|x)}[\mathcal{L}(x,y)] - \mathcal{H}(q_{\psi}(y|x)) = \mathcal{U}(x)
\end{split}
\end{equation}
% \begin{equation}
%     q_{\phi, \psi}(y, z|x) = q_{\phi}(z|x, y)q_{\psi}(y|x) 
% \end{equation}
where $q_{\phi}(z|x, y)$ and $q_{\psi}(y|x)$ are inference models for latent variable $z$ and $y$ respectively, which have a factorised form $q_{\phi, \psi}(y, z|x) = q_{\phi}(z|x, y)q_{\psi}(y|x)$; 
$\mathcal{H}(\cdot)$ denotes causal entropy;
$\mathcal{L}(x, y)$ is the variational bound for labeled data, ans is formulated as:
\begin{equation}
\begin{split}
    \mathcal{L}(x, y) = \mathbb{E}_{q_{\phi}(z|x,y)}[p_{\theta}(x|z,y)] +\log p(y)\\ 
    -\text{KL}(q_{\phi}(z|x,y)||p(z))
\end{split}
\end{equation}
where KL is the Kullback-Leibler divergence, and  $p(y)$, $p(z)$ are the prior distribution of $y$, $z$.

The generative model $p_{\theta} $, inference model $q_{\phi}$ and $q_{\psi}$ are optimized using both the labeled subset $p_l$ and unlabeled subset $p_u$ using the objective as:
\begin{equation}
    \mathcal{L} = \sum_{(x,y)\sim p_l}\mathcal{L}(x,y) + \sum_{x \sim p_u} \mathcal{U}(x)
\end{equation}
% where $p_l$ and $p_u$ denote labeled and unlabeled subsets, respectively.

% [approaches to semi-supervised label prediction, and the basic components of this framework]
\subsubsection*{Semi-Supervised Action Prediction} 
We now describe the learning of action prediction model $f_{\mathcal{A}}$ using semi-supervised expert demonstrations.
We extend the semi-supervised VAE by modeling the generation process of \emph{state transitions}. 
State transition information is indicative for action prediction and is available in both fully and partially labeled demonstrations.
Thus we choose to describe the generation process of state transitions, and the optimization objective is formulated as: 
% Thus we directly formulate the optimization objective as state tracking task:
\begin{equation} \label{act-generation}
\begin{split}
\log &p_{\theta}(s_{t+1},s_{t}) = \log \sum_{a} \int p_{\theta}(s_{t+1},z, s_t, a) dz \\
    & = \log \sum_{a} \int p(s_{t+1}, s_{t}|,z, a)p(z)p(a)dz
\end{split}
\end{equation}
% Note that we also make all the inferences depending on system utterance $\vect{u}_t$, but we omit this for simplicity.
For partially labeled dialogues, we treat action labels as latent variables and use the action prediction model $f_{\mathcal{A}}(a|u_t, s_{t}, s_{t+1})$ to infer the value (which is denoted as $f_{\mathcal{A}}(a|\cdot)$ later for simplicity).     
The variational bound of Eqn. \ref{act-generation} is derived as:
% \log p_{\theta}(s_{t+1},s_{t}, a_t) \geq
\begin{equation}
\begin{split}
     \mathcal{U}(s_{t+1},s_t) = \mathbb{E}_{f_{\mathcal{A}}(a|\cdot)}[\mathcal{L}(s_{t+1},s_t, a)] \\-\mathcal{H}(f_{\mathcal{A}}(a|\cdot)) 
\end{split}
\end{equation}
where $\mathcal{L}(s_{t+1},s_t, a_t)$ is the variational bound for demonstrations with action labels and is derived as:
\begin{equation}
\begin{split}
    \mathcal{L}(s_{t+1},s_t, a) = \mathbb{E}_{q_{\phi}(z|u_t,a)}[p_{\theta}(s_{t+1}|s_{t}, z)] \\ - \text{KL}(q_{\phi}(z|u_t,a)||p(z))
\end{split}
\end{equation}
where $q_{\phi}(z|u_t,a)$ is the inference model for latent variable $z$.
% [benefits of such modelling: state transition is essential and more indicative for action prediction, and is easier to capture compared textual system utterance ]
% [related state transitions used to be applied as feature, we now use it as objective]
Lastly, we use fully annotated samples to form a classification loss:
\begin{equation}
   \mathcal{L}_{\text{cls}}  = \mathbb{E}_{\tau_i \in D_{\mathcal{F}}} [\log f_{\mathcal{A}}(a|u_t,s_t,s_{t+1})]
\end{equation}
% (s_t,x_t,s_{t+1},a_t)

The overall objectives includes the loss of fully and partially labeled demonstrations:
\begin{equation} \label{act-overall}
\begin{aligned}
    \mathcal{L}_{\text{act}} = & \sum_{\tau_i \in D_{\mathcal{F}} } \mathcal{L}(s_{t+1}, s_t, a) + \\ &\sum_{\tau_i \in D_{\mathcal{P}} } \mathcal{U}_(s_{t+1}, s_t) + \mathcal{L}_{\text{cls}}
\end{aligned}
\end{equation}

\subsubsection*{Action Embeddings Learning}
We then incorporate action embedding function $f_E$ into the developed semi-supervised action prediction approach.
The reason to introduce action embeddings is to make the learning of reward estimator more efficient and robust.
Specifically, prediction error of the action prediction model might impinge the learning of reward estimator, especially for our semi-supervised scenarios where fully labeled dialogues are limited.
% thus can  and provide better generalization over actions for reward estimation 
% The predicted system action might not exactly match the true label, 
By mapping actions to an embedding space, `wrongly predicted' partially labeled demonstrations can still provide sufficient knowledge and thus we could achieve better generalization over actions for reward estimation.

To this aim, we consider the inference steps in the semi-supervised learning process and utilize the ones that involve action labels, i.e., the inference models for latent variables $z$ and $a$.
We first specify how the action prediction model is modified to include action embeddings.
Inspired by \cite{chandak2019learning}, we model the action selection using Boltzmann distribution for stability during training:  
\begin{equation} \label{g-function}
\begin{aligned}
    &f_{\mathcal{A}}(a|u_t, s_{t}, s_{t+1})= \frac{e^{z_a/\gamma}}{\sum_{a' \in \mathcal{A}} e^{z_{a'}/\gamma}}\\
     &z_a = e(a)^{\top} g(u_t, s_{t}, s_{t+1}), e(a)=f_{E}(a)
\end{aligned}
\end{equation}
where $\gamma$ is a temperature parameter, and $g(\cdot)$ is a function that maps the input into hidden states of the same dimension as action embeddings.
% , and can be implemented in various ways, e.g., MLP.
We also modify the inference model for latent variable by incorporating action embeddings:
\begin{equation}
    q_{\phi}(z|u_t,a) =  q_{\phi}(z|u_t,e(a))
\end{equation}

% what we obtain
After optimizing the action prediction model $f_{\mathcal{A}} $ and action embedding function $f_E$ jointly using the objective function Eqn. \ref{act-overall}, we use action embeddings to enrich the expert demonstrations.
For fully labeled dialogues, we map the given system action labels to corresponding embeddings and obtain $\tau_i=\{(s_t, e(a_t))|1 \leq t \leq n_{\tau} \}$.
For partially labeled dialogues, we first infer the action using prediction model: $\Tilde{a}_t=f_{\mathcal{A}}(u_t, s_t,s_{t+1}) $, and map the inferred action to its embedding to obtain: $\tau_j = \{(s_t, e(\Tilde{a}_t))|1 \leq t \leq n_{\tau}  \} $.

\subsection{Reward Estimation by Dynamics Model} \label{reward-estimation}

% (should be highlighted) reason to estimate reward based on the dialogue progress
We aim to learn a reward estimator based on action representations obtained from the action learning module.
To achieve a more stable reward estimation than adversarial reward learning, we propose a reward estimator based on \emph{dialogues progress}.
% what is dialogue progress
Dialogue progress describes how user goals are achieved through multistep interactions and can be modeled as dialogue state transitions.
We argue that an action should be given higher rewards when it leads to similar dialogue progress (i.e., state transitions) of expert demonstrations.
To this aim, we learn a model to explicitly model dialogue progress without the negative sampling required by adversarial learning, and rewards can be estimated as the local-probabilities assigned to the taken actions. 

% take advantage of the expert demonstrations enriched by action embeddings and

% we use VRNN to model dialogue progress, a few reasons to do so
To model dialogue progress, we use variational recurrent neural network (VRNN) \cite{chung2015recurrent}.
The reason to use a stochastic dynamics model is due to the `one-to-many' nature of task-oriented dialogues.
Specifically, both user and dialogue system have multiple feasible options to proceed the dialogues which requires the modeling of uncertainty. 
% which cause difficulty for a deterministic sequential model to well capture dialogue progress.
Thus, by adding latent random variables to an RNN architecture, VRNN can provide better modeling of dialogue progress than deterministic dialogue state tracking.

% components (stochastic, deterministic, observations) and (prior, inference, generation, recurrence)
VRNN has three types of variables: the observations (and here we consider action embeddings), the stochastic state $z$, and the deterministic hidden state $h$, which summarizes previous stochastic states $z_{\leq t}$, and previous observations $a_{\leq t}$. 
We formulate the prior stochastic states to be conditioned on previous timesteps through hidden state $h_{t-1}$:
% and dialogue state label $s_{t-1}$ :
\begin{equation}
    p(z_t|a_{<t}, z_{<t})=\varphi^{\text{prior}}(h_{t-1})
\end{equation}
We obtain posterior stochastic states by incorporating the observation at the current step, i,e. action embeddings $e(a_t)$:
\begin{equation}
    q(z_t|a_{\leq t},z_{<t}) = \varphi^{\text{enc}}(h_{t-1}, e(a_t))
\end{equation}
Predictions are made by decoding latent states, including both the stochastic and deterministic:
\begin{equation}\label{VRNN-dec}
    p(e(a_t)|z_{\leq t}, a_{<t}) = \varphi^{\text{dec}}(z_t, h_{t-1}, s_t)
\end{equation}
And lastly the deterministic states are updated as:
\begin{equation}
    h_t = \varphi^{\text{rnn}}(e(a_t), z_t, h_{t-1}, s_t)
\end{equation}
where $\varphi$ are all implemented as neural networks. 
Note that we also make the prediction and recurrence step to condition on the dialogue state $s_t$ to provide more information.

% optimization objective
We train the VRNN by optimizing the evidence lower bound (ELBO) as:
\begin{equation}
\begin{aligned}
    \mathcal{L}&_{\text{VRNN}} = \mathbb{E}_{q(z_t|a_{\leq t}, z_{<t } )} \big[ \sum_{t} \log p(e(a_t)|z_{\leq t}, a_{<t})  \\ &- \text{KL}(q(z_t|a_{\leq t}, z_{<t } )||p(z_t|a_{< t}, z_{<t } )) \big] 
\end{aligned}
\end{equation}
% reward estimation using learned VRNN
% policy learning using VRNN based reward estimator (and to clarify that various approaches can be used)  
% With the learned VRNN, we set the reward function as:
The rewards are estimated as the conditional probability given the hidden state of VRNN, which encodes the current dialogue progress:  
\begin{equation}
    r(s_{\leq t}, a_t)= \log p_{\varphi^{\text{dec}}}(a_t|a_{<t} ,s_{\leq t})
\end{equation}
where $p_{\varphi^{\text{dec}}}$ is the probability given to the selected action based on the decoding step of VRNN (Eqn. \ref{VRNN-dec}).
% The reward estimation in this way goes beyond state-action pairs by taking the dialogue progress into account and, and thus can lead to better generalization abilities.
The larger this conditional probability is, the more similar the dialogue progress this action leads to imitates the expert demonstrations.
The proposed reward estimation is agnostic to the choice of policy, and various approaches (e.g., Deep Q-learning, Actor-Critic) can be optimized by plugging into the policy learning objective (Eqn. \ref{policy-learning}).

\subsection{Expanding to Unlabeled Corpus} \label{expand-unlabel}
% reasons to consider unlabeled corpus, and notations for fully and unlabeled dialogues
We further describe how to expand the proposed model, including action learning and reward estimation modules, to utilize \emph{unlabeled expert demonstrations}. 
Formally, we consider the setting that we have fully labeled dialogues $D_{\mathcal{F}}$ and unlabeled dialogues $D_{\mathcal{U}}$. 
For each dialogue in $D_{\mathcal{U}}$, only textual conversations are provided and neither of state and action labels are available: $\tau_j = \{ (c_t, u_t)|1 \leq t \leq n_{\tau} \} $, where $c_t$ is the context and consists of the dialogue history of both user and system utterances.

% how we formulate the action prediction model 
With the absence of dialogue state information, we formulate the action prediction model as 
$f_{\mathcal{A}}(a|u_t,u_{t-1},u_{t+1})$.
This formulation can be considered as an application of Skip-Thought \cite{kiros2015skip}, which originally utilizes contextual sentences as supervision signals.
In our scenarios, we instead utilize the previous and next system utterances to provide more indicative information for action prediction.
% The reason to include the previous and next system utterances, $u_{t-1} $ and $u_{t+1} $, is that such sequential information is indicative for action prediction 
% and is also relatively easier to model since they all follow expert policies. 

% generation model for semi-supervised action prediction under unlabeled data setting
We also build the joint learning of action prediction model the action embeddings on semi-supervised VAE framework.
Instead of modeling state transitions, we choose the process of \emph{response generation} to fully utilize unlabeled dialogues:
\begin{equation}
\begin{split}
    \log p_{\theta}(u_t) = \log \sum_{a} \int p_{\theta}(u_t, z, a)dz \\
     = \log \sum_{a} \int p_{\theta}(u_t|z, a_t)p(z)p(a)dz
\end{split}    
\end{equation}
System action labels are treated as latent variables for unlabeled dialogues, and the variational bond is derived as: 
\begin{equation}\label{expand-unsupervised}
\begin{split}
    \mathcal{U}(u_t)=\mathbb{E}_{f_{\mathcal{A}}(a|\cdot)}[\mathcal{L}(u_t,a)]-\mathcal{H}(f_{\mathcal{A}}(a|\cdot))
\end{split}
\end{equation}
where $\mathcal{L}(u_t,a)$ is variational bound for fully labeled dialogues:
\begin{equation}
\begin{split}\label{expand-supervised}
    \mathcal{L}(u_t,a) = \mathbb{E}_{q_{\phi}(z|u_t, a)}[p_{\theta}(u_t|z, u_{t-1}, u_{t+1} ) ] \\
    -\text{KL}({q_{\phi}(z|a, u_t)}||p(z))
\end{split}
\end{equation}

The objective to jointly train the prediction model and action embeddings is the same as Eqn. \ref{act-overall}, where the terms for fully and partially labeled dialogues are replaced with the ones in Eqn. \ref{expand-supervised} and \ref{expand-unsupervised}, respectively.
% setting of all types of dialogues
Such expanding also enables a sufficient semi-supervised learning when expert demonstrations include all types of labeled dialogues: $D_{\mathcal{F}}$, $D_{\mathcal{P}}$ and $D_{\mathcal{U}}$.
We notice that the posterior approximation $q_{\phi}(z|u_t, a) $ and action embedding function $f_E $ can be sharing between the process of state transitions and response generation. 
Thus, by treating semi-supervised learning in $D_{\mathcal{F}}$ and $D_{\mathcal{P}}$ as auxiliary constraints, the learning over unlabeled corpus can also benefit from dialogues state information.

\section{Experiments}

\begin{table*} [tbp]
    \centering
\begin{threeparttable}
\caption{ Semi-Supervised Policy Learning Results ($D_{\mathcal{F}} $ and $D_{\mathcal{P}} $) 
} \label{exp-F+P}
% \tiny
% \scriptsize
% \footnotesize
\small
% \normalsize  
\setlength\tabcolsep{3.5pt}
\begin{tabular}{l|l|ccc|ccc|ccc}
\toprule
% Method & nDCG & $\alpha\mbox{-}$nDCG & p$\mbox{-}$nDCG & nDCG & $\alpha\mbox{-}$nDCG & p$\mbox{-}$nDCG \\
\multicolumn{2}{l|}{} & \multicolumn{3}{c|}{$D_{\mathcal{F}}$(5\%) + $D_{\mathcal{P}}$(95\%)} & \multicolumn{3}{c|}{$D_{\mathcal{F}}$(10\%) + $D_{\mathcal{P}}$(90\%)} &  \multicolumn{3}{c}{$D_{\mathcal{F}}$(20\%) + $D_{\mathcal{P}}$(80\%)} \\
\cmidrule{3-11}
\multicolumn{2}{c|}{ \textsc{Model}} & Entity-F1 & Success  &Turns & Entity-F1 & Success &Turns & Entity-F1 & Success  &Turns  \\
\midrule
Handcrafted & PPO & 41.8& 34.1 &13.3 & 45.3&36.7&12.5 & 50.6&41.2&11.2 \\
\midrule
\multirow{2}{*}{\makecell[l]{Reward \\Learning}  } & ALDM & 38.7&35.6&15.2&42.1&38.6&14.9&44.9&42.1&13.7\\
& GDPL &49.5&47.5&12.8&54.9&53.2&12.1&60.4&59.1&10.8  \\
\midrule
\multirow{3}{*}{\makecell[l]{Semi-VAE\\Enhanced  } }& SS-PPO  &45.2&36.2&13.6&47.4&37.2&12.4&53.1&43.6&11.5 \\
& SS-ALDM &39.6&38.8&14.7&44.7&43.8&13.2&47.8&51.3&12.4  \\
& SS-GDPL &53.7&51.2&11.1&61.3&58.4&10.5&66.5&68.7&9.2\\
\midrule
\multirow{3}{*}{Proposed} & SS-VRNN  &68.7&63.2&9.4&75.1&68.5&8.6&77.3&72.4&8.2\\
& Act-GDPL &  70.6&65.6&9.5&78.8&71.1&8.4&80.9&78.0&8.2 \\
& Act-VRNN & \textbf{76.2}&\textbf{72.7}&\textbf{9.1}&\textbf{83.0}&\textbf{81.8}&\textbf{8.0}&\textbf{85.5}&\textbf{86.7}&\textbf{7.9}\\
\bottomrule
\end{tabular}
\end{threeparttable}
\end{table*}

% settings
To show the effectiveness of the proposed model (denoted as \textbf{Act-VRNN}), we experiment on a multi-domain dialogue environment under semi-supervised setting (Sec. \ref{exp-settings}).
% overall results
We compare against state-of-the-art approaches, and their variants enhanced by semi-supervised learning techniques (Sec. \ref{exp-results}).
% with and without a combination with semi-supervised learning techniques (Sec. 5.2).
% discussions
We analyze the effectiveness of action learning and reward estimation of Act-VRNN under different supervision ratios (Sec. \ref{exp-discussins}). 

\subsection{Settings} \label{exp-settings}
We use MultiWOZ \cite{budzianowski2018multiwoz}, a multi-domain human-human conversational dataset in our experiments.
It contains in total 8438 dialogues spanning over seven domains, and each dialogue has 13.7 turns on average.
% difficult for task-oriented policy learning compared to movie-for
MultiWOZ also contains a larger dialogue state and action space compared to former datasets such as movie-ticket booking dialogues \cite{li2017end}, and thus it is a much more challenging environment for policy learning. 
% methods to utilize MultiWOZ for policy learning: building user simulator, 
To use MultiWOZ for policy learning, a user simulator that initializes a user goal at the beginning and interacts with dialogue policy is required. 
For a fair comparison, we adopt the same procedure as Takanobu et al.~(\citeyear{takanobu2019guided}) to train the user simulator based on auxiliary user action annotations provided by ConvLab \cite{lee2019convlab}.

% setting of semi-supervised policy learning 
To simulate semi-supervised policy learning, we remove system action and dialogue states annotations to obtain partially labeled and unlabeled expert demonstrations, respectively.
% Note that [since , can be regarded as expert demonstrations  ].
Fully labeled expert demonstrations are randomly sampled from all training dialogues with different ratios (5\%, 10\%, and 15\% in our experiments).   
% We conduct experiments at each ratio on five different sampling results and maintain the distribution of dialogues over domains at each sampling.
Note that the absence of action or state annotations only applies for expert demonstrations, while interactions between policy and user simulator are in dialogue-act level as \cite{takanobu2019guided} and not affected by semi-supervised setting.

We use a three-layer transformer \cite{vaswani2017attention} with a hidden size of 128 and 4 heads as our base model for action embedding learning, i.e., $g(\cdot)$ in Eqn. \ref{g-function}.
We use grid search to find the best hyperparameters for the models. 
We choose the action embedding dimensionality among \{50, 75, 100, 150, 200\}, the stochastic latent state size in VRNN among \{16, 32, 64, 128, 256\}, and the deterministic latent state size among \{25, 50, 75, 100, 150\}.

% evaluation metrics
We use \textbf{Entity-F1} and \textbf{Success Rate} to evaluate dialogue task completion.
Entity-F1 computes the F1 score based on whether the requested information and indicated constraints from users are satisfied.
Compared to inform rate and match rate used by Budzianowski et al.~(\citeyear{budzianowski2018multiwoz}), Entity-F1 considers both informed and requested entities at the same time and balances the recall and precision. 
Success rate indicates the ratio of successful dialogues, where a dialogue is regarded as successful only if  all informed and requested entities are matched of the dialogue. 
We use \textbf{Turns} to evaluate the cost for task completion, where a lower number indicates the policy performs tasks more efficiently.

% baselines and variants of Act-VRNN to compare with
We compare Act-VRNN with three policy learning baselines:
(1) \textbf{PPO} \cite{schulman2017proximal} using hand-crafted rewards setting;
% and only uses expert demonstrations for policy pre-training;
(2) \textbf{ALDM}\cite{liu2018adversarial};
(3) \textbf{GDPL} \cite{takanobu2019guided};
% baseline variants
We further consider using semi-supervised techniques to enhance the baselines under semi-supervised setting, and denote them as \textbf{SS-PPO}, \textbf{SS-ALDM}, and \textbf{SS-GDPL}.
%   add footnote to clarify SSVAE achieves the best prediction accuracies among other approaches, bootstrap
Specifically, we first train a prediction model based on semi-supervised VAE \cite{kingma2014semi}, and use the prediction results as action annotations for expert demonstrations.~\footnote{We also experimented with the pseudo-label approach \cite{lee2013pseudo}, and the empirical results were worse than Semi-VAE. Thus, we only report the Semi-VAE enhancement results in the table for simplicity.}
% Note that semi-supervised enhancing demonstrations also benefit PPO
% Act-VRNN variants
We also compare the full model \textbf{Act-VRNN} with its two variants:
(1) \textbf{SS-VRNN} uses a VRNN that consumes predicted action labels instead of action embeddings;
(2) \textbf{Act-GDPL} feeds expert demonstrations enriched by action embeddings to the same reward function as GDPL
% , which is then trained in a adversarial way as \cite{fu2017learning} .

\subsection{Overall Results} \label{exp-results}
% {$D_{\mathcal{F}}$(10\%) +\\$D_{\mathcal{U}}$
% F+P
Table \ref{exp-F+P} shows that our proposed model consistently outperforms other models in the setting that uses fully and partially annotated dialogues ($D_{\mathcal{F}}$ and $D_{\mathcal{P}}$).
Act-VRNN improves task completion (measured by Entity-F1 and Success) while requiring less cost (measured by Turns).
%   example at primary results
For example, Act-VRNN (81.8) outperforms SS-GDPL (60.4) by 35.4\% under Success when having 10\% fully annotated dialogues, and requires the fewest turns.
%   Act-VRNN benefit from components: both action learning and  are 
Meanwhile, we find that both action learning and dynamics model are essential to the superiority of Act-VRNN.
For example, Act-VRNN achieves 19.8\% and 11.2\% improvements over SS-VRNN and Act-GDPL, respectively, under Success when having 20\% fully annotated dialogues. 
% the ratio of $D_{\mathcal{F}}$ is 20\%.
This validates that the learned action embeddings well capture similarities among actions, and VRNN is able to exploit such similarities for reward estimation.  
%   insights from comparing baselines: reward learning is important, benefits from semi-supervised techniques is limited especially when ratio of fully annotated dialogues is low 

We further find that the improvements brought by semi-VAE enhancement is limited for baselines, especially when the ratio of fully annotated dialogues is low.
% without using semi-VAE
For example, SS-PPO and SS-GDPL achieve 6\% and 7\% improvements over their counterparts under Success when having 5\% fully annotated dialogues.
Similar results are also observed for pseudo-label approach.
In general, the pseudo-label methods are outperformed by the counterparts of Semi-VAE and are even worse than the baselines without enhancement when the ratio of fully annotated dialogues is low. 
For example, in setting $D_{\mathcal{F}}+D_{\mathcal{P}}$, pseudo-label enhanced PPO performs worse than PPO under Entity-F1 when the ratio of fully annotated dialogues is 5\% and 10\% (37.2 vs 41.8, 39.2 vs 45.3), and only achieves slightly gain when the ratio is 20\% (51.0 vs 50.6).
This is largely because the prediction accuracy of Semi-VAE and pseudo-label approach might be low with a small amount of fully annotated dialogues, and the expert dialogues with mispredicted actions impinge reward function learning of baselines.  
Act-VRNN overcomes this challenge with the generalization ability brought by modeling dialogue progress in an action embedding space for reward estimation.

% \vspace{-0.9pt}
% F+U and F+U+P
The results for policy learning using unlabeled dialogues ($D_{\mathcal{U}}$) are shown on Table \ref{exp-F+U}.
We consider two settings: (1) having fully labeled and unlabeled dialogues, i.e., $D_{\mathcal{F}}$ + $D_{\mathcal{U}}$; (2) having all three types of dialogues , i.e., $D_{\mathcal{F}}$ + $D_{\mathcal{P}}$ +$D_{\mathcal{U}}$. 
We can see that Act-VRNN significantly outperforms the baselines in both settings.
For example, in setting $D_{\mathcal{F}}$ + $D_{\mathcal{U}}$, Act-VRNN outperforms SS-GDPL by 43\% and 44\% under Entity-F1 and Success, respectively. 
Similar results are also observed in setting $D_{\mathcal{F}}$ + $D_{\mathcal{P}}$ +$D_{\mathcal{U}}$.
%   relative advantage of SS-VRNN and Act-GDPL, and left for the following section 
We further find that SS-VRNN outperforms Act-GDPL in these two settings while the results are opposite in setting $D_{\mathcal{F}}$ + $D_{\mathcal{P}}$, and we will conduct a detailed discussion in the following section.
%   benefits from partially labeled  
By comparing results of Act-VRNN and baselines in these two settings, we can see that Act-VRNN can better exploit the additional partially labeled dialogues.
For example, SS-GDPL only achieves 2.3\% under Success while Act-VRNN achieves more than 5\%. 

\hspace{0.3pt}

\subsection{Discussions} \label{exp-discussins}
\begin{figure}[t]
\centering
\subfigure[\small{$D_{\mathcal{F}}$(5\%) + $D_{\mathcal{P}}$(95\%) }]{
\begin{overpic}[height=3.40cm]{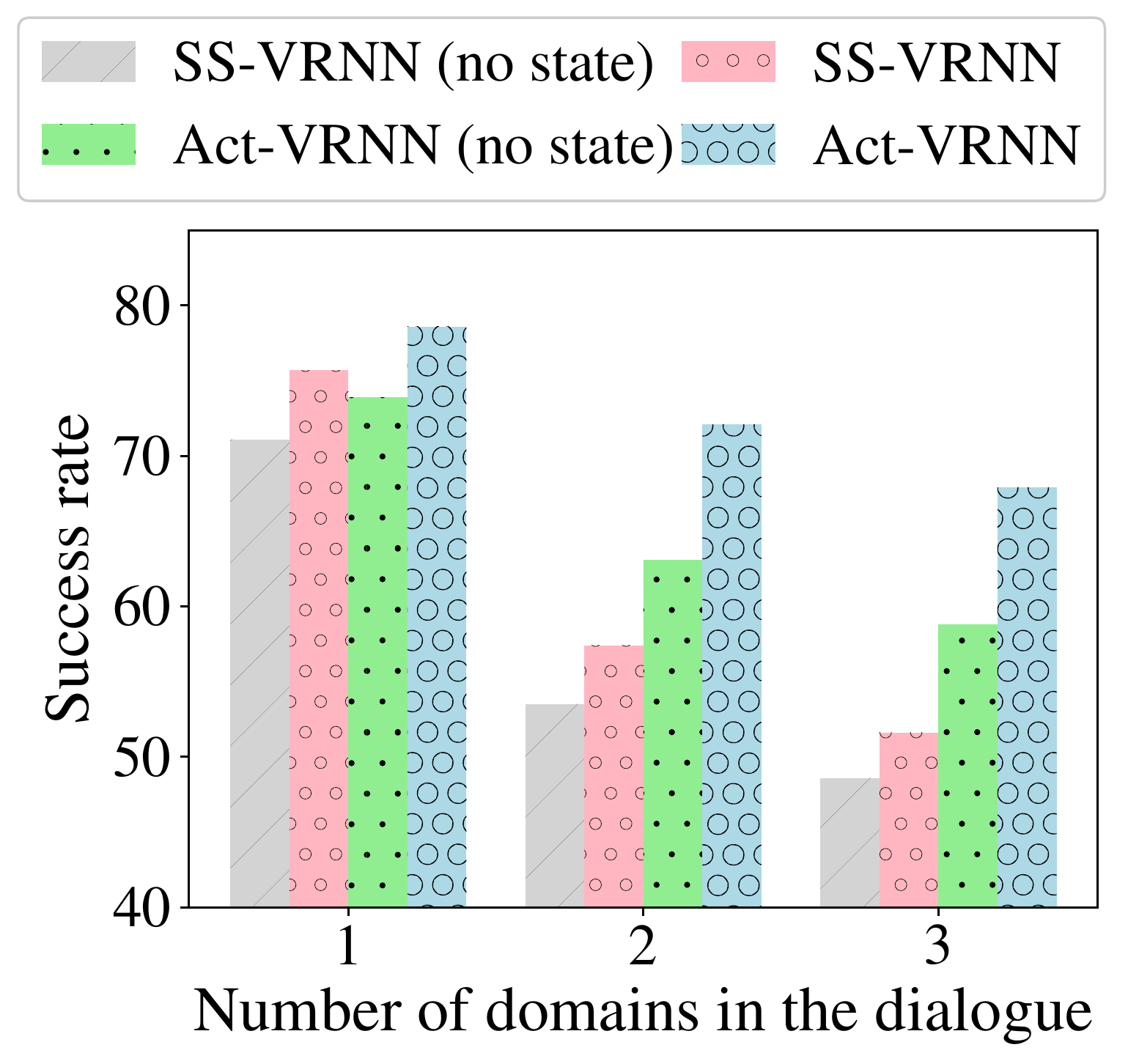}
\label{act-F5}
\end{overpic}
}
\subfigure[\small{$D_{\mathcal{F}}$(20\%) + $D_{\mathcal{P}}$(80\%)}]{
\begin{overpic}[height=3.40cm]{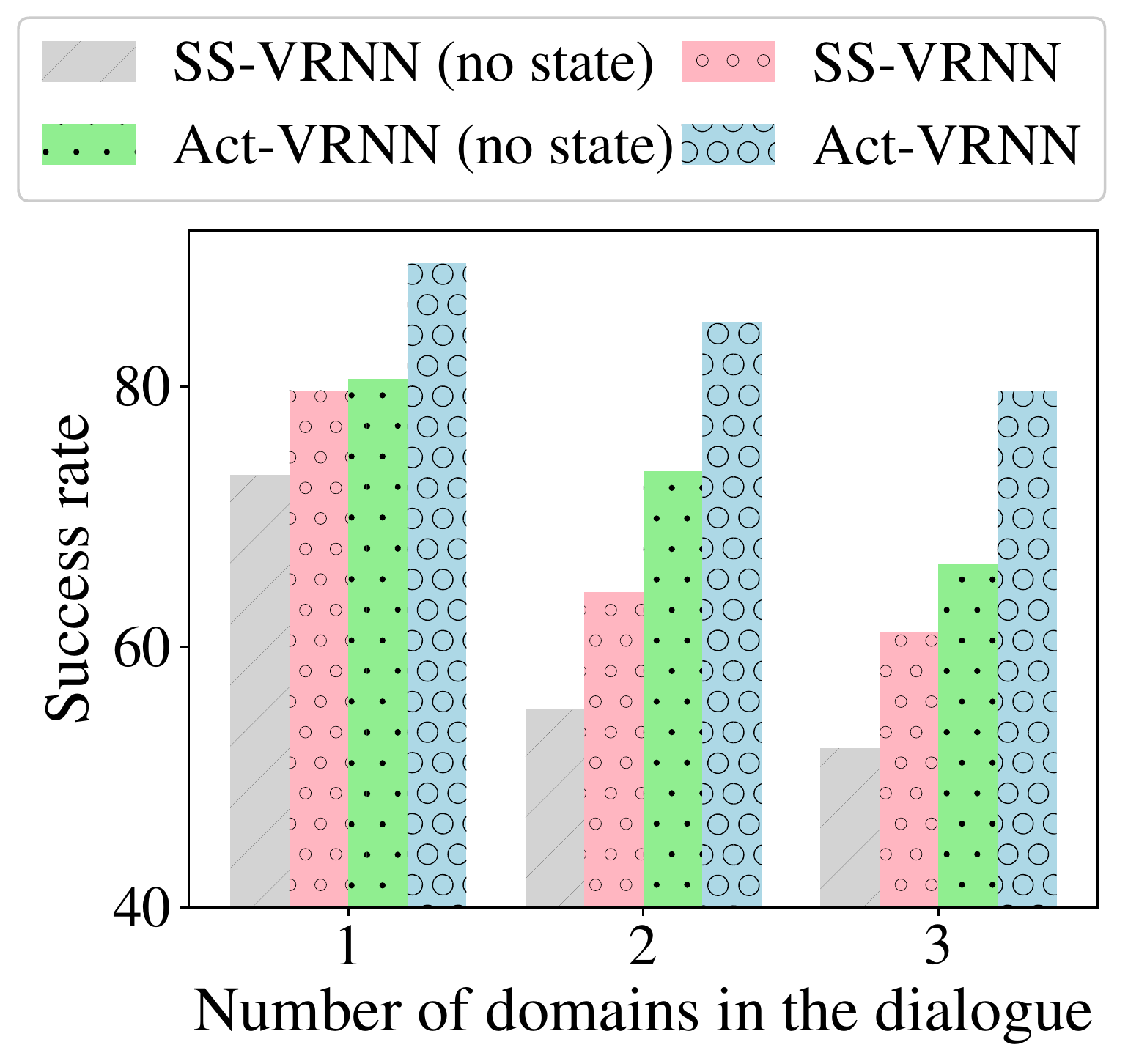}
\label{act-F20}
\end{overpic}
}
\caption{Effects of action learning ($D_{\mathcal{F}} $ and $D_{\mathcal{P}}$)}
\label{effect-act}
\end{figure}

% act feedings
We first study the effects of action learning module in Act-VRNN.
%   exp setting
We compare Act-VRNN with SS-VRNN, and their counterparts that do not use state transition based objective in semi-supervised learning (i.e., optimizing Eqn. 3 instead of Eqn. 7).
These two variants are denoted as Act-VRNN (no state) and SS-VRNN (no state).
%   domain
For a thorough investigation, under each setting, we further show the performances under dialogues spanning over different number of domains. 
Dialogues spanning over more domains are considered more difficult. 
%   results
The results under two supervision ratio setting are shown in Fig. \ref{act-F5} and Fig. \ref{act-F20}.
%   Act-VRNN better
We can see that Act-VRNN outperforms other variants in each configuration, especially in the dialogues that include more than one domains.
This is largely because the learned action embeddings effectively discover the similarities between actions across domains, and thus lead to better generalization of reward estimation.      
%   Act stable than SS
We further find that the state transition based objective we formulated fits well with the VRNN based reward estimator.
Both Act-VRNN and SS-VRNN optimized considering state transitions achieve performance gains.

\begin{table}[tbp]
    \centering
\begin{threeparttable}
\caption{ Semi-Supervised Policy Learning Results ($D_{\mathcal{F}} $, $D_{\mathcal{P}} $, and $D_{\mathcal{U}} $) 
} \label{exp-F+U}
% \tiny
% \scriptsize
% \footnotesize
\small
\setlength\tabcolsep{4.0pt}
\begin{tabular}{l|l|ccc}
\toprule
% Method & nDCG & $\alpha\mbox{-}$nDCG & p$\mbox{-}$nDCG & nDCG & $\alpha\mbox{-}$nDCG & p$\mbox{-}$nDCG \\
\multicolumn{1}{c|}{ \textsc{Supervision}} & \multicolumn{1}{c|}{ \textsc{Model}} & Entity-F1 & Success  &Turns  \\
\midrule
\multirow{7}{*}{\makecell[l]{$D_{\mathcal{F}}$(10\%) +\\$D_{\mathcal{U}}$(90\%)}} & ALDM & 40.0& 34.9& 15.9 \\
\cmidrule{2-5}
& SS-PPO  & 44.7& 33.8& 12.9 \\
& SS-ALDM & 42.1& 36.4& 14.9 \\
& SS-GDPL & 56.3& 50.2& 11.8\\
\cmidrule{2-5}
& SS-VRNN  & 74.1& 67.1& 9.1\\
& Act-GDPL & 72.9& 66.7& 8.5\\
& Act-VRNN & \textbf{80.6}& \textbf{72.4}& \textbf{8.4}\\
\midrule
\midrule
\multirow{7}{*}{\makecell[l]{$D_{\mathcal{F}}$(10\%) +\\$D_{\mathcal{P}}$(10\%) + \\$D_{\mathcal{U}}$(80\%)}} & ALDM & 41.7& 35.2& 15.7\\
\cmidrule{2-5}
& SS-PPO  &  44.9& 34.6& 12.8\\
& SS-ALDM &  42.5& 40.1& 14.7\\
& SS-GDPL& 57.1&51.4& 10.7 \\
\cmidrule{2-5}
& SS-VRNN&   75.6& 67.9& 8.8\\
& Act-GDPL & 73.3& 67.1& 8.5\\
& Act-VRNN & \textbf{81.1}& \textbf{76.3}& \textbf{8.2}\\
\bottomrule
\end{tabular}
\begin{tablenotes}\footnotesize
\item[*] Note that PPO and GDPL achieve the same results as $D_{\mathcal{F}}$(10\%)+$D_{\mathcal{P}}$(90\%) in Table \ref{exp-F+P} since they can only utilize dialogues in $D_{\mathcal{F}}$
\end{tablenotes}
\end{threeparttable}
\end{table}

\begin{figure}[!t]
\centering
\subfigure[\small{$D_{\mathcal{F}}$ + $D_{\mathcal{P}}$ }]{
\begin{overpic}[height=3.28cm]{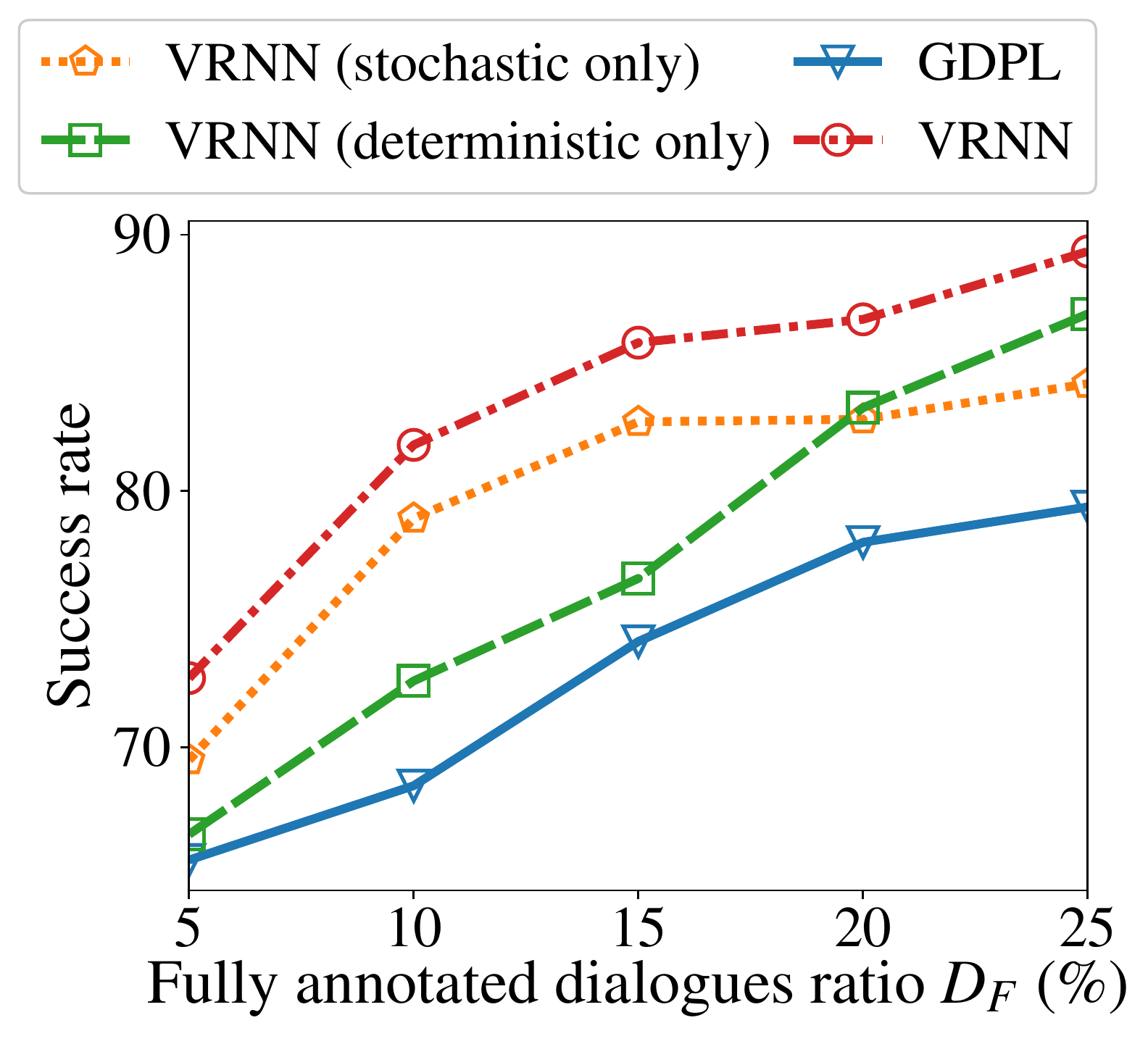}
\label{vrnn-FP}
\end{overpic}
}
\subfigure[\small{$D_{\mathcal{F}}$ + $D_{\mathcal{U}}$}]{
\begin{overpic}[height=3.28cm]{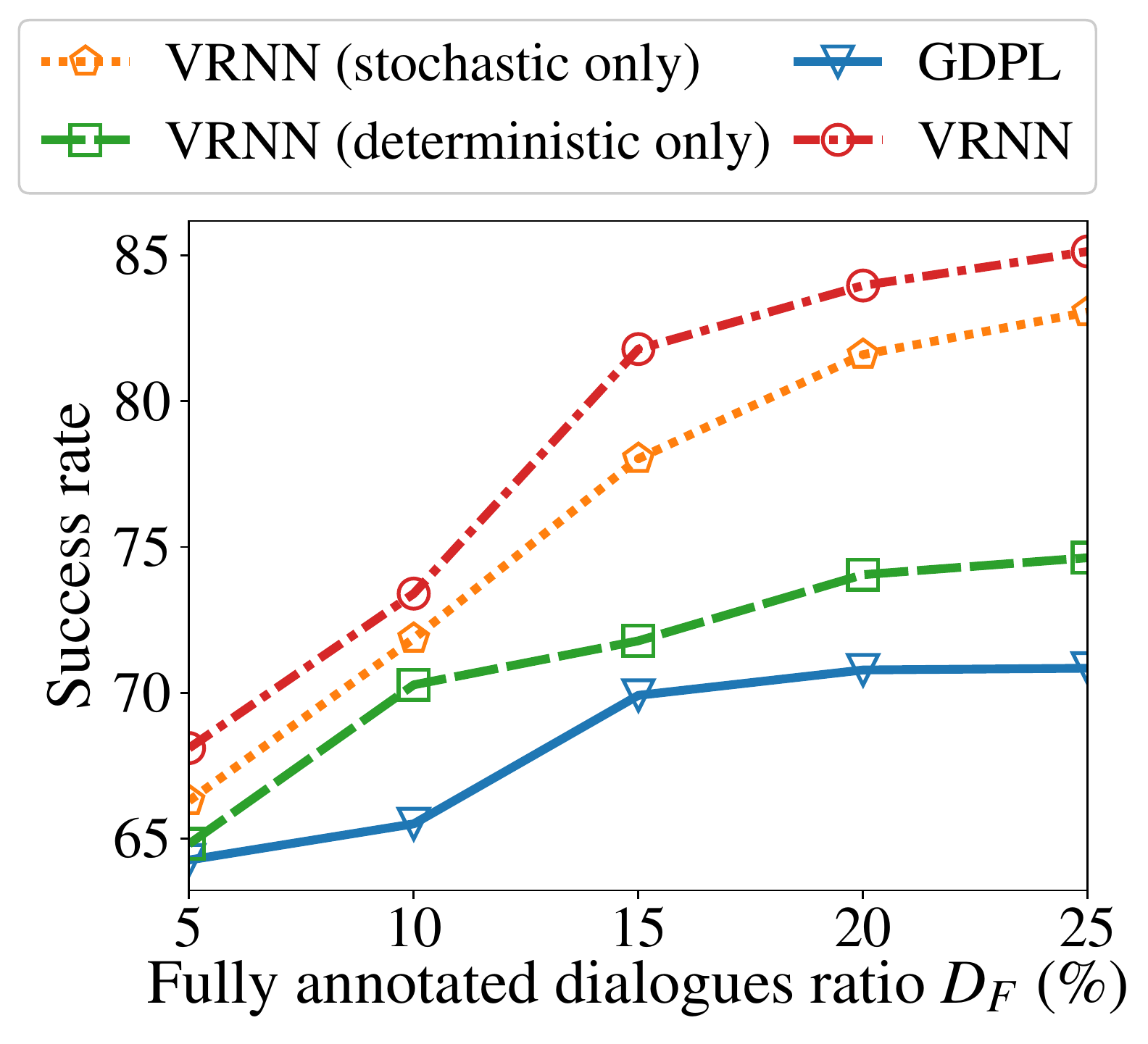}
\label{vrnn-FU}
\end{overpic}
}
\caption{Effects of dynamics model}
\label{effect-vrnn}
\end{figure}

% VRNN variants
Last, we study the effects of dynamics model based reward function in Act-VRNN.
%   exp setting
We consider four different models as reward function: (1) our full dynamics model VRNN; (2) a dynamics model having only deterministic states (Eqn. 17); (3) a dynamics model having only stochastic states (Eqn. 15); (4) GDPL.
All four models are learned based on action embedding learned in the action learning module.
The results under $D_{\mathcal{F}}$ + $D_{\mathcal{P}}$ and $D_{\mathcal{F}}$ + $D_{\mathcal{U}}$ are shown in Fig. \ref{vrnn-FP} and Fig. \ref{vrnn-FU}, respectively.
%   both important
We can see that both stochastic and deterministic states in VRNN are important, since VRNN outperforms its two variants and GDPL in each configuration.
% For example, VRNN outperforms its two variants and GDPL in $D_{\mathcal{F}}$ + $D_{\mathcal{P}}$
%   further find that the contribution of stochastic and deterministic states may vary in different setting
We further find that the contribution of stochastic and deterministic states may vary in different setting.
For example, VRNN (stochastic only) consistently outperforms VRNN (deterministic only) in $D_{\mathcal{F}}$ + $D_{\mathcal{U}}$ while opposite results are observed in $D_{\mathcal{F}}$ + $D_{\mathcal{P}}$ when ratio of $D_{\mathcal{F}}$ is over 20\%.
This is largely because modeling dialogue progress using stochastic states can provide more stable with less supervision signals, while the incorporation of deterministic can lead to more precise estimation can when more information of expert demonstrations are available.

\section{Related Work}
% \subsubsection*{Reward Learning for Dialogue Policy}
Reward learning aims to provide more effective and sufficient supervision signals for dialogue policy.
Early studies focus on learning reward function utilizing external evaluations, e.g., user experience feedbacks \cite{gavsic2013line}, objective ratings \cite{su2015reward,ultes2017reward}, or a combination of multiple evaluations \cite{su2016line,Chen2019MiningUR}.
These approaches often assume a human-in-the-loop setting where interactions with real users are available during training, which is expensive and difficult to scale.
As more large-scale high-quality dialogue corpus become available (e.g., MultiWOZ \cite{budzianowski2018multiwoz}), recent years have seen a growing interest in learning reward function from expert demonstrations.
Most recent approaches apply inverse reinforcement learning techniques for dialogue policy learning \cite{takanobu2019guided,li2019dialogue}. 
These all require a complete state-action annotation for expert demonstrations.
We aim to overcome this limitation in this study.

% Semi-supervised learning for text classification aims to utilize unlabeled data to boost model performance.
Semi-supervised learning aims to utilize unlabeled data to boost model performance, and is studied in computer vision \cite{Iscen2019LabelPF}, item ranking \cite{Park2019AdversarialSA,huang2019carl}, and multi-label classification \cite{miyato2015distributional,wang2018kdgan,wang2019adversarial}.
Many studies apply semi-supervised VAE \cite{kingma2014semi} for different classification tasks, e.g., sentiment analysis \cite{xu2017variational,li-etal-2019-semi-supervised}, text matching \cite{shen2018deconvolutional,choi-etal-2019-cross}.
While these work focus on prediction accuracies, we aim to enrich expert demonstrations via semi-supervised learning.

% \subsubsection*{Semi-supervised learning for text classification}

\section{Conclusions}
We study the problem of semi-supervised policy learning and propose Act-VRNN to provide more effective and stable rewards estimations. 
We formulate a generative model to jointly infer action labels and learn action embeddings.
We design a novel reward function to first model dialogue progress, and estimate action rewards by determining whether the action leads to similar progress as expert dialogues.
The experimental results confirm that Act-VRNN achieves better task completion compared with the state-of-the-art in two settings that consider partially labeled or unlabeled dialogues.
% potential of policy learning for
For future work, we will explore the scenarios that annotations are absent for all expert dialogues.

\section*{Acknowledgement}
We would like to thank Xiaojie Wang for his help.
This work is supported by Australian Research Council (ARC) Discovery Project DP180102050,
and China Scholarship Council (CSC).

\bibliography{acl2020.bib}
\bibliographystyle{acl_natbib}

\end{document}